\documentclass{amsart}

\usepackage{amsmath}
\usepackage{amsthm}
\usepackage{amsfonts}
\usepackage{amssymb}
\usepackage{graphicx}
\usepackage{xcolor}
\usepackage[hyphens,spaces,obeyspaces]{url}
\usepackage{hyperref}
\usepackage{subcaption}
\usepackage{etoolbox}
\usepackage{enumitem}
\usepackage[foot]{amsaddr}
\usepackage{etoolbox}
\usepackage{multicol}
\usepackage{tcolorbox}
\usepackage{datetime}
    \newdateformat{monthyeardate}{%
  \monthname[\THEMONTH], \THEYEAR}
\usepackage{setspace}
\usepackage{longtable}
\usepackage[authoryear,compress]{natbib}
\usepackage{tikz}
\usetikzlibrary{calc}
\usetikzlibrary{decorations.pathreplacing}
\usetikzlibrary{patterns}
\usepackage{array}
\usepackage{epigraph}
\hypersetup{
    colorlinks,
    linkcolor={red!50!black},
    citecolor={blue!50!black},
    urlcolor={blue!80!black}
}

\makeatletter
    \newtheoremstyle{indented}
        {12pt}% space before
        {12pt}% space after
        {\addtolength{\@totalleftmargin}{3.5em}
        \addtolength{\linewidth}{-3.5em}
        \parshape 1 3.5em \linewidth}% body font
        {}% indent
        {\bfseries}% header font
        {:}% punctuation
            {.5em}% after theorem header
        {}% header specification (empty for default)
\makeatother

\makeatletter
    \newtheoremstyle{indentedProp}
        {12pt}% space before
        {12pt}% space after
        {\addtolength{\@totalleftmargin}{3.5em}
        \addtolength{\linewidth}{-3.5em}
        \parshape 1 3.5em \linewidth}% body font
        {}% indent
        {\bfseries}% header font
        {:}% punctuation
            {0.5em}% after theorem header
        {}% header specification (empty for default)
\makeatother

\theoremstyle{indented}
    
\theoremstyle{indentedProp}
    
\theoremstyle{indented}
    
\theoremstyle{indented}
    
\theoremstyle{indented}
    
\theoremstyle{indented}
    
\theoremstyle{indented}

\makeatletter
    \patchcmd{\NAT@test}{\else \NAT@nm}{\else \NAT@nmfmt{\NAT@nm}}{}{}
    \DeclareRobustCommand\citepos
        {\begingroup
        \let\NAT@nmfmt\NAT@posfmt% ...except with a different name format
        \NAT@swafalse\let\NAT@ctype\z@\NAT@partrue
        \@ifstar{\NAT@fulltrue\NAT@citetp}{\NAT@fullfalse\NAT@citetp}}
    \let\NAT@orig@nmfmt\NAT@nmfmt
    \def\NAT@posfmt#1{\NAT@orig@nmfmt{#1's}}
\makeatother

\title{Deep Learning and Ethics}

\author{Travis LaCroix$^1$ \\ Simon J. D. Prince$^2$}
\address{$^1$Department of Philosophy \\ Dalhousie University}
\email{$^1$tlacroix@dal.ca}

\address{$^2$Department of Computer Science \\University of Bath}

\date{Unpublished draft of \monthyeardate\today. This article appears as chapter 21 of \citet{Prince-2023-UDL}; a complete draft of the textbook is available here: \href{http://udlbook.com}{http://udlbook.com}. \\ Copyright in this Work has been licensed exclusively to The MIT Press, \href{https://mitpress.mit.edu}{https://mitpress.mit.edu}, which will be releasing the final version to the public in 2023. All inquiries regarding rights should be addressed to The MIT Press, Rights and Permissions Department.}

\begin{document}

\maketitle

    AI is poised to change society for better or worse.  These technologies have enormous potential for social good \citep{taddeo2018ai,tomavsev2020ai}, including important roles in healthcare \citep{rajpurkar2022ai} and the fight against climate change \citep{rolnick2023tackling}.  However, they also have the potential for misuse and unintended harm.  This has led to the emergence of the field of {\em AI ethics}.

    The modern era of deep learning started in 2012 with AlexNet, but sustained interest in AI ethics did not follow immediately. Indeed, a workshop on fairness in machine learning was rejected from NeurIPS 2013 for want of material.   It wasn't until 2016 that AI Ethics had its ``AlexNet'' moment, with  ProPublica's expos{\'e} on bias in the COMPAS recidivism-prediction model \citep{Angwin-et-al-2016} and Cathy O'Neil's book {\it Weapons of Math Destruction} \citep{ONeil-2016}.  Interest has swelled ever since; submissions to the Conference on {\it Fairness, Accountability, and Transparency} (FAccT) have increased nearly ten-fold in the five years since its inception in 2018.

    In parallel, many organizations have proposed policy recommendations for responsible AI. \citet{Jobin-et-al-2019} found $84$ documents containing AI ethics principles, with $88$\% released since 2016. This proliferation of non-legislative policy agreements, which depend on voluntary, non-binding cooperation, calls into question their efficacy \citep{McNamara-2018, Hagendorff-2019, LaCroix-Mohseni-2022}.    In short, AI Ethics is in its infancy, and ethical considerations are often reactive rather than proactive.

    This chapter considers potential harms arising from the design and use of AI systems.  These include algorithmic bias, lack of explainability, data privacy violations, militarization, fraud, and environmental concerns.   The aim is not to provide advice on being more ethical.  Instead, the goal is to express ideas and start conversations in key areas that have received attention in philosophy, political science, and the broader social sciences.

\section{Value alignment} \label{sec:ethics_VAP}

    When we design AI systems,  we wish to ensure that their ``values'' (objectives) are aligned with those of humanity. This is sometimes called the {\it value alignment problem} \citep{Russell-2019, Christian-2020, Gabriel-2020}.  This is challenging for three reasons.  First, it's difficult to define our values completely and correctly.  Second, it is hard to encode these values as objectives of an AI model, and third, it is hard to ensure that the model learns to carry out these objectives.

    In a machine learning model, the loss function is a proxy for our true objectives, and a misalignment between the two is termed the {\em outer alignment problem} \citep{Hubinger-et-al-2019}.   To the extent that this proxy is inadequate,  there will be ``loopholes'' that the system can exploit to minimize its loss function while failing to satisfy the intended objective.  For example, consider training an RL agent to play chess. If the agent is rewarded for capturing pieces, this may result in many drawn games rather than the desired behavior (to win the game).  In contrast, the {\em inner alignment problem} is to ensure that  the behavior of an AI system does not diverge from the intended objectives even when the loss function is well specified.  If the learning algorithm fails to find the global minimum or the training data are unrepresentative, training can converge to a  solution that is misaligned with the true objective resulting in undesirable behavior \citep{Goldberg-1987, Mitchell-et-al-1992, Lehman-Stanley-2008}.

    \citet{Gabriel-2020} divides the value alignment problem  into {\em technical} and {\em normative} components.   The technical component concerns how we encode values into the models so that they reliably do what they should.  Some concrete problems, such as avoiding reward hacking and safe exploration, may have purely technical solutions \citep{Amodei-et-al-2016}.   In contrast, the normative component concerns what the correct values are in the first place.  There may be no single answer to this question, given the range of things that different cultures and societies value.  It's important that the encoded values are representative of everyone and not just culturally dominant subsets of society.

\begin{figure}
        \begin{center}
            \includegraphics[width=1.0\linewidth]{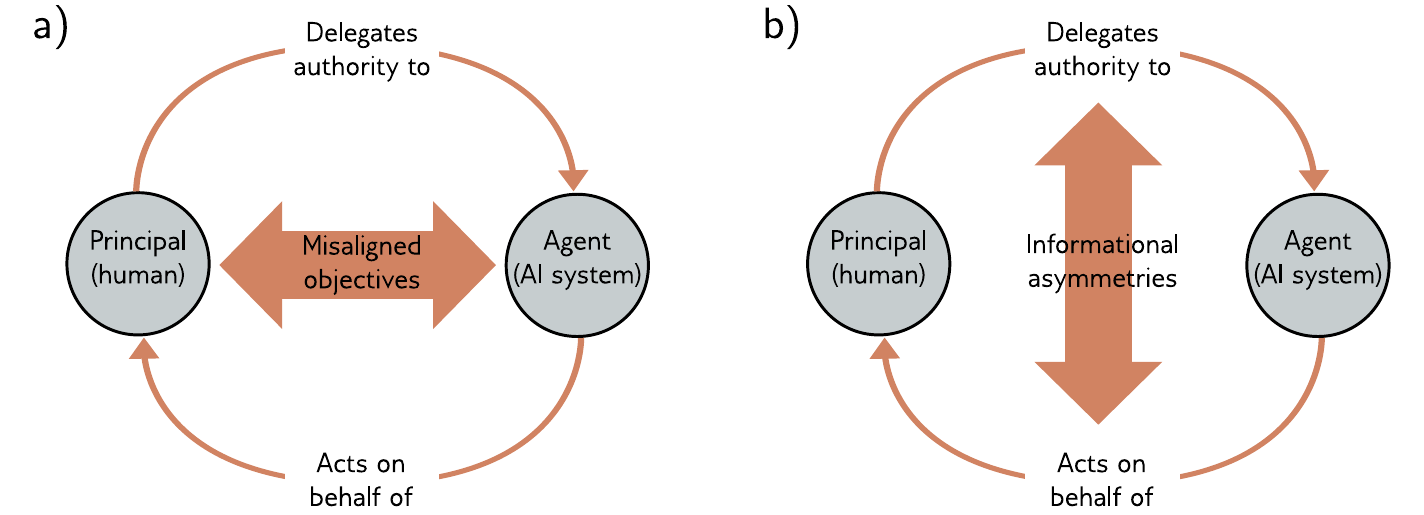}
        \caption{\footnotesize Structural description of the value alignment problem. a) Problems arise from a) misaligned objectives (e.g., bias) or b) informational asymmetries between a (human) principal and an (artificial) agent (e.g., lack of explainability). Adapted from \protect\citet{LaCroix2023}.}
        \label{fig:ethics_structural}
        \end{center}
    \end{figure}

    Another way to think about value alignment is as a {\it structural} problem that arises when a human {\em principal} delegates tasks to an artificial {\em agent} \citep{LaCroix-2022-Linguistic}.   This is similar to the principal-agent problem in economics \citep{Laffont-Martimort-2002}, which allows that there are competing incentives inherent in any relationship where one party is expected to act in another's best interests. In the AI context, such conflicts of interest can arise when either (i) the objectives are misspecified or (ii) there is an informational asymmetry between the principal and the agent (figure~\ref{fig:ethics_structural}).

    Many topics in AI ethics can be understood in terms of this structural view of value alignment.  The following sections discuss problems of bias and fairness and artificial moral agency (both pertaining to specifying objectives) and transparency and explainability (both related to informational asymmetry).

\subsection{Bias and fairness}\label{sec:ethics_Bias}

    From a purely scientific perspective, bias refers to statistical deviation from some norm. In AI, it can be pernicious when this deviation depends on {\it illegitimate} factors that impact an output. For example, gender is irrelevant to job performance, so it is illegitimate to use gender as a basis for hiring a candidate. Similarly, race is irrelevant to criminality, so it is illegitimate to use race as a feature for recidivism prediction.

    Bias in AI models can be introduced in various ways \citep{Fazelpour-Danks-2021}:

    \begin{itemize} 
        \setlength \itemsep{-0.0em}
        \item {\bf Problem specification:}  Choosing a model's goals requires a value judgment about what is important to us, which allows for the creation of biases \citep{Fazelpour-Danks-2021}. Further biases may emerge if we fail to operationalize these choices successfully and the problem {\it specification} fails to capture our intended goals \citep{Mitchell-et-al-2021}.
        
        \item {\bf Data:}  Algorithmic bias can result when the dataset is unrepresentative or incomplete \citep{Danks-London-2017}.  For example, the PULSE face super-resolution algorithm \citep{menon2020pulse} was trained on a database of photos of predominantly white celebrities.  When applied to a low-resolution portrait of Barack Obama, it generated a photo of a white man \citep{vincent2020what}.
        
        If the society in which training data are generated is structurally biased against marginalized communities, even complete and representative datasets will elicit biases \citep{Mayson-2018}.  For example, Black individuals in the US have been policed and jailed more frequently than white individuals. Hence, historical data used to train recidivism prediction models are already biased against Black communities.  
        \item {\bf Modeling and validation:} Choosing a mathematical definition to measure model fairness requires a value judgment.  There exist distinct but equally-intuitive definitions that are logically inconsistent \citep{Kleinberg-et-al-2016, Chouldechova-2016, Berk-et-al-2017}. This suggests the need to move from a purely mathematical conceptualization of fairness toward a more substantive evaluation of whether algorithms promote justice in practice \citep{Green-2022}. 
    
        \item {\bf Deployment:} Deployed algorithms may interact with other algorithms, structures, or institutions in society to create complex feedback loops that entrench extant biases \citep{ONeil-2016}. For example, large language models like GPT3 \citep{brown2020language} are trained on web data. However, when GPT3 outputs are published online, the training data for future models is degraded.   This may exacerbate biases and generate novel societal harm \citep{Falbo-LaCroix-2022}. 
    \end{itemize}

    Unfairness can be exacerbated by considerations of {\it intersectionality}; social categories can combine to create overlapping and interdependent systems of oppression. For example, the discrimination experienced by a queer woman of color is not merely the sum of the discrimination she might experience as queer, as gendered, or as racialized \citep{Crenshaw-1991}.  Within AI, \citet{Buolamwini-Gebru-2018} showed that face analysis algorithms trained primarily on lighter-skinned faces underperform for darker-skinned faces.  However, they perform even worse on combinations of features such as skin color and gender than might be expected by considering those features independently.

    Of course, steps can be taken to ensure that data are diverse, representative, and complete.  But if the society in which the training data are generated is structurally biased against marginalized communities, even completely accurate datasets will elicit biases. In light of the potential for algorithmic bias and the lack of representation in training datasets described above, it is also necessary to consider how failure rates for the outputs of these systems are likely to exacerbate discrimination against already-marginalized communities \citep{Buolamwini-Gebru-2018, Raji-Buolamwini-2019, Raji-et-al-2022}. The resulting models may codify and entrench systems of power and oppression, including capitalism and classism; sexism, misogyny, and patriarchy; colonialism and imperialism; racism and white supremacy; ableism; and cis- and heteronormativity.   A perspective on bias that maintains sensitivity to power dynamics requires accounting for historical inequities and labor conditions encoded in data \citep{Micelli-et-al-2022}.

    \begin{figure}[t]
        \definecolor{cellColor}{rgb}{0.75,0.75,0.75}
        \footnotesize
        \begin{center}
        %\ra{1.1}
        \resizebox{\textwidth}{!}{
        \begin{tabular}{llll}
        \hline 
         {Data collection } & {Pre-processing} & {Training} & {Post-processing} \\
         
        \hline 
        \parbox[t][][t]{3cm}{$\bullet$ Identify lack of\\{\color{white} $\bullet$} examples or\\{\color{white} $\bullet$} variates and collect}  &\parbox[t][][t]{2.6cm}{ $\bullet$ Modify labels\\$\bullet$ Modify input data\\ $\bullet$ Modify input/\\{\color{white} $\bullet$} output pairs} & \parbox[t][][t]{3.4cm}{ $\bullet$ Adversarial training\\$\bullet$ Regularize for fairness\\$\bullet$ Constrain to be fair} &\parbox[t][][t]{3cm}{$\bullet$ Change thresholds\\$\bullet$ Trade-off accuracy\\{\color{white} $\bullet$} for fairness}\\
        \hline
        \end{tabular}}\normalsize
        \caption{ \footnotesize Bias mitigation.  Methods have been proposed to compensate for bias at all stages of the training pipeline,
        from data collection to post-processing of already trained models.  See \protect\citet{Barocas-et-al-2019} and \protect\citet{mehrabi2022survey}. }\label{fig:ethics_fairness}
        \end{center}
    \end{figure}

    To prevent this, we must actively ensure that our algorithms are fair.  A na\"ive approach is {\em fairness through unawareness} which simply removes the {\em protected attributes} (e.g., race, gender) from the input features.   Unfortunately, this is ineffective; the remaining features can still carry information about the protected attributes.  More practical approaches first define a mathematical criterion for fairness.  For example, the {\em separation} measure in binary classification requires that the prediction $\hat{y}$ is conditionally independent of the protected variable $a$ (e.g., race) given the true label $y$. %in contrast, the {\em sufficiency} measure requires that the true label $y$ is independent of the protected variable $a$ given the prediction $\hat{y}$.  
    Then they intervene in various ways to minimize the deviation from this measure (figure~\ref{fig:ethics_fairness}).

    A further complicating factor is that we cannot tell if an algorithm is unfair to a community or take steps to avoid this unless we can establish community membership.  Most research on algorithmic bias and fairness has focused on ostensibly {\it observable} features that might be present in training data (e.g., gender). However, features of marginalized communities may be {\it unobservable}, making bias mitigation even more difficult. Examples include queerness \citep{Tomasev-et-al-2021}, disability status, neurotype, class, and religion.   A similar problem occurs when observable features have been excised from the training data to prevent models from exploiting them.

\subsection{Artificial moral agency}

    Many decision spaces do not include actions that carry moral weight. For example, choosing the next chess move has no obvious moral consequence. However, elsewhere actions can carry moral weight. Examples include decision-making in autonomous vehicles \citep{Awad-et-al-2018, Evans-et-al-2022}, lethal autonomous weapons systems \citep{Arkin-2008a, Arkin-2008b}, and professional service robots for childcare, elderly care, and health care \citep{Anderson-Anderson-2008, Sharkey-Sharkey-2012}.  As these systems become more autonomous, they may need to make moral decisions independent of human input.

    This leads to the notion of {\it artificial moral agency}. An artificial moral agent is an autonomous AI system capable of making moral judgments. Moral agency can be categorized in terms of increasing complexity \citep{Moor-2006}:

    \begin{enumerate}
    \setlength\itemsep{-0.3em}
        %%%%%%%%%%
        \item {\bf Ethical impact agents} are agents whose actions have ethical impacts. Hence, almost any technology deployed in society might count as an ethical impact agent. 
        %%%%%%%%%%
        \item {\bf Implicit ethical agents} are ethical impact agents that include some in-built safety features. 
        %%%%%%%%%%
        \item {\bf Explicit ethical agents} can contextually follow general moral principles or rules of ethical conduct. 
        %%%%%%%%%%
        \item {\bf Full ethical agents} are agents with beliefs, desires, intentions, free will, and consciousness of their actions.
        %%%%%%%%%%
    \end{enumerate}

    The field of machine ethics seeks approaches to creating artificial moral agents. These approaches can be categorized as {\em top-down}, {\em bottom-up}, or {\em hybrid} \citep{Allen-et-al-2005}. Top-down (theory-driven) methods directly implement and hierarchically arrange concrete rules  based on some moral theory to guide ethical behavior.  Asimov's ``Three Laws of Robotics'' are a trivial example of this approach.

    In bottom-up (learning-driven) approaches, a model learns moral regularities from data without explicit programming \citep{Wallach-et-al-2008}.  For example, \citet{Noothigattu-et-al-2018} designed a voting-based system for ethical decision-making that uses data collected from human preferences in moral dilemmas to learn social preferences; the system then summarizes and aggregates the results to render an ``ethical'' decision. Hybrid approaches combine top-down and bottom-up approaches.

    Some researchers have questioned the very idea of artificial moral agency and argued that moral agency is unnecessary for ensuring safety \citep{Van-Wynsberghe-Robbins-2019}. See \citet{Cervantes-et-al-2019} for a recent survey of artificial moral agency and \citet{Tolmeijer-et-al-2020} for a recent survey on technical approaches to artificial moral agency.

\subsection{Transparency and opacity}

    A complex computational system is {\em transparent} if all of the details of its operation are known.  A system is {\em explainable} if humans can understand how it makes decisions.  In the absence of transparency or explainability, there is an asymmetry of information between the user and the AI system, which makes it hard to ensure value alignment.

    \citet{Creel-2020} characterizes transparency at several levels of granularity.  {\em Functional transparency} refers to knowledge of the algorithmic functioning of the system (i.e., the logical rules that map inputs to outputs).  The methods in this book are described at this level of detail.  {\em Structural transparency} entails knowing {\it how} a program executes the algorithm.  This can be obscured when commands written in high-level programming languages are executed by machine code. Finally, {\em run transparency} requires understanding how a program was executed in a particular instance.  For deep networks, this includes knowledge about the hardware, input data, training data, and interactions thereof.  None of these can be ascertained by scrutinizing code.

    For example, GPT3 is functionally transparent;  its architecture is described in \citet{brown2020language}. However, it does not exhibit structural transparency as we do not have access to the code, and it does not exhibit run transparency as we have no access to the learned parameters, hardware, or training data.   The subsequent version GPT4 is not transparent at all.  The details of how this commercial product works are unknown.

\subsection{Explainability and interpretability}

    \begin{figure}
    \begin{center}
    \includegraphics[width=0.9\linewidth]{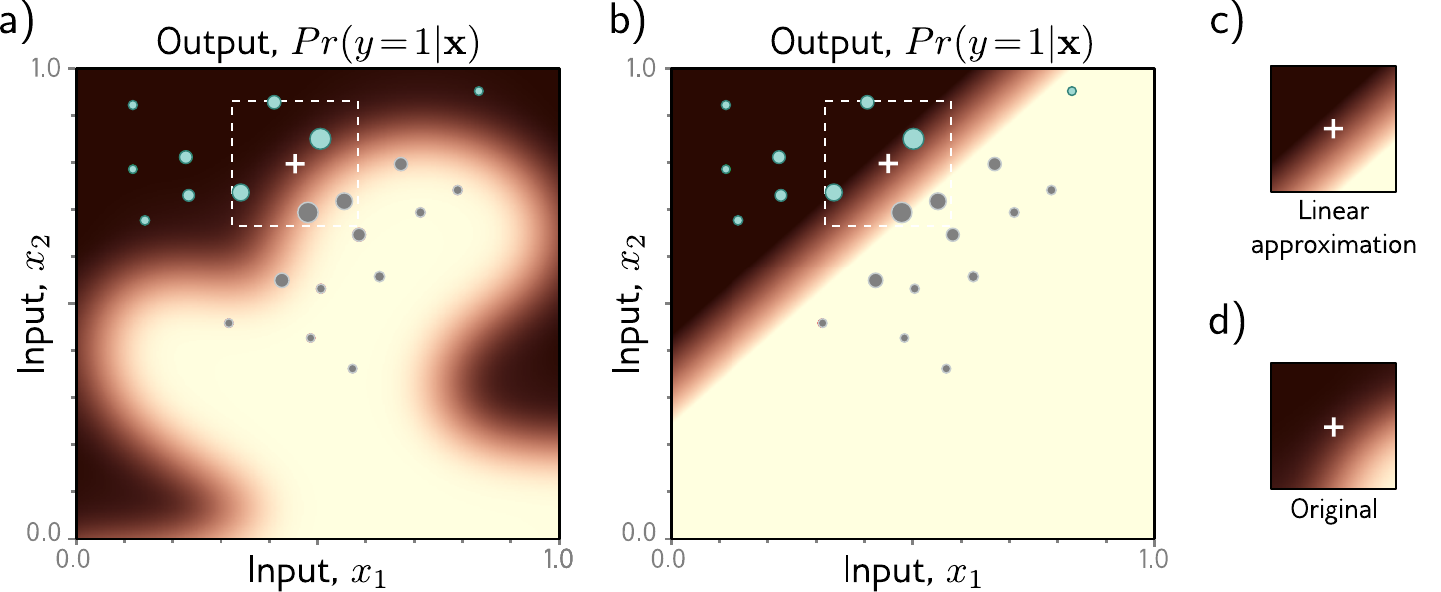}
    \caption{\footnotesize{LIME. Output functions of deep networks are complex; in high dimensions, it's hard to know why a decision was made or how to modify the inputs to change it without access to the model.  a) Consider trying to understand why $Pr(y=1|\bf{x})$ is low at the white cross. LIME probes the network at nearby points to see if it identifies these as $Pr(y=1|\bf{x})<0.5$ (cyan points) or $Pr(y=1|\bf{x})\geq0.5$ (gray points).  It weights these points by proximity to the point of interest (weight indicated by circle size).  b) The weighted points are used to train a simpler model (here, logistic regression --- a linear function passed through a sigmoid).  c) Near the white cross, this approximation is close to d) the original function. Even though we did not have access to the original model, we can deduce from the parameters of this approximate model, that if we increase $x_1$ or decrease $x_2$, $Pr(y=1|\bf{x})$ will increase, and the output class will change. Adapted from \cite{prince2022explain}.}    }\label{fig:ethics_lime}
    \end{center}
    \end{figure}

    Even if a system is  transparent, this does not imply that we can understand how a decision is made or what information this decision is based on.  Deep networks may contain billions of parameters, so there is no way we can understand how they work based on examination alone.   However, in some jurisdictions, the public may have a right to an explanation.  Article 22 of the EU General Data Protection Regulation suggests all data subjects should have the right to ``obtain an explanation of the decision reached'' in cases where a decision is based solely on automated processes.\footnote{Whether Article 22 {\em actually} mandates such a right is debatable \cite[see][]{Wachter-et-al-2017}.}

    These difficulties have led to the sub-field of explainable AI. One moderately successful area is producing local explanations. Although we can't explain the entire system, we can sometimes describe how a particular input was classified. For example, {\em Local interpretable model-agnostic explanations} or {\em LIME} \citep{Ribeiro-et-al-2016} samples the model output at nearby inputs and uses these samples to construct a simpler model (figure~\ref{fig:ethics_lime}).  This provides insight into the classification decision, even if the original model is neither transparent nor explainable.

    It remains to be seen whether it is possible to  build complex decision-making systems that are fully understandable to their users or even their creators.  There is also an ongoing debate about what it means for a system to be explainable, understandable, or interpretable \citep{Erasmus-et-al-2021}; there is currently no concrete definition of these concepts. See \citet{molnar2020interpretable} for more information.

\section{Intentional misuse}\label{sec:ethics_Misuse}

    The problems in the previous section arise from poorly specified objectives and informational asymmetries. However, even when a system functions correctly, it can entail unethical behavior or be intentionally misused.  This section highlights some specific ethical concerns arising from the misuse of AI systems.

\subsection{Face recognition and analysis}

    Face recognition technologies have an especially high risk for misuse. Authoritarian states can use them to identify and silence protesters, thus risking democratic ideals of free speech and the right to protest. \citet{Smith-Miller-2022} argue that there is a mismatch between the values of liberal democracy (e.g., security, privacy, autonomy, and accountability) and the potential use cases for these technologies (e.g., border security, criminal investigation and policing, national security, and  the commercialization of personal data). Thus, some researchers, activists, and policymakers have questioned whether this technology should exist \citep{Barrett-2020}.

    Moreover, these technologies often do not do what they purport to \citep{Raji-et-al-2022}. For example, the New York Metropolitan Transportation Authority moved forward with and expanded its use of facial recognition despite a proof-of-concept trial reporting a 100\% failure rate to detect faces within acceptable parameters \citep{Berger-2019}.  Similarly, facial analysis tools often oversell their abilities \citep{Raji-Fried-2021}, dubiously claiming to be able to infer individuals' sexual orientation \citep{Leuner-2019}, emotions \citep{Stark-Hoey-2021}, hireability \citep{Fetscherin-et-al-2020}, or criminality \citep{Wu-Zhang-2016}. \citet{Stark-Hutson-2022} highlight that computer vision systems have created a resurgence in the ``scientifically baseless, racist, and discredited pseudoscientific fields'' of physiognomy and phrenology.

\subsection{Militarization and political interference}

    Governments have a vested interest in funding AI research in the name of national security and state building. This risks an arms race between nation-states, which carries with it ``high rates of investment, a lack of transparency, mutual suspicion and fear, and a perceived intent to deploy first'' \citep{Sisson-et-al-2020}.

    Lethal autonomous weapons systems receive significant attention because they are easy to imagine, and indeed many such systems are under development \citep{heikkilla2022business}. However, AI also facilitates cyber-attacks and disinformation campaigns (i.e., inaccurate or misleading information that is shared with the intent to deceive). AI systems allow the creation of highly realistic fake content and facilitate the dissemination of information, often to targeted audiences \citep{Akers-et-al-2018} and at scale \citep{Bontridder-Poullet-2021}.

    \citet{Kosinski-et-al-2013} suggest that sensitive variables, including sexual orientation, ethnicity, religious and political views, personality traits, intelligence, happiness, use of addictive substances, parental separation, age, and gender, can be predicted by ``likes'' on social media alone. From this information, personality traits like ``openness'' can be used for manipulative purposes (e.g., to change voting behavior).

\subsection{Fraud}

    Unfortunately, AI is a useful tool for automating fraudulent activities (e.g., sending mass emails or text messages that trick people into revealing sensitive information or sending money). Generative AI can be used to deceive people into thinking they are interacting with a legitimate entity or generate fake documents that mislead or deceive people. Additionally, AI could increase the sophistication of cyber-attacks, such as by generating more convincing phishing emails or adapting to the defenses of targeted organizations.

    This highlights the downside of calls for transparency in machine learning systems: the more open and transparent these systems are, the more vulnerable they may be to security risks or use by bad-faith actors. For example, generative language models, like ChatGPT, have been used to write software and emails that could be used for espionage, ransomware, and other malware \citep{Goodin-2023}.

    The tendency to anthropomorphize computer behaviors and particularly the projection of meaning onto strings of symbols is termed the {\em ELIZA effect} \citep{Hofstadter-1995}.  This leads to a false sense of security when interacting with sophisticated chatbots, making people more susceptible to text-based fraud such as  romance scams or business email compromise schemes \citep{Abrahams-2023}. \citet{Veliz-2023} highlights how emoji use in some chatbots is inherently manipulative, exploiting instinctual responses to emotive images.

\subsection{Data privacy}

    Modern deep learning methods rely on huge crowd-sourced datasets, which may contain sensitive or private information. Even when sensitive information is removed, auxiliary knowledge and redundant encodings can be used to de-anonymize datasets \citep{Narayanan-Shmatikov-2008}. Indeed, this famously happened to the Governor of Massachusetts, William Weld, in 1997. After an insurance group released health records that had been stripped of obvious personal information like patient name and address, an aspiring graduate student was able to ``de-anonymize'' which records belonged to Governor Weld by cross-referencing with public voter rolls.

    Hence, privacy-first design is important for ensuring the security of individuals' information, especially when applying deep learning techniques to high-risk areas such as healthcare and finance. Differential privacy and semantic security (homomorphic encryption or secure multi-party computation) methods can be used to ensure data security during model training \cite[see][]{Mireshghallah-et-al-2020, Boulemtafes-et-al-2020}.

\section{Other social, ethical, and professional issues}\label{sec:ethics_Other}

    The previous section identified areas where AI can be deliberately misused.  This section describes other potential side effects of the widespread adoption of AI.

\subsection{Intellectual property}

    Intellectual property (IP) can be characterized as non-physical property that is the product of original thought \citep{Moore-Himma-2022}.   In practice, many AI models are trained on copyrighted material. Consequently, these models' deployment can pose legal and ethical risks and run afoul of intellectual property rights \citep{Henderson-et-al-2023}.

    Sometimes, these issues are explicit.  When language models are prompted with excerpts of copyrighted material, their outputs may include copyrighted text verbatim, and similar issues apply in the context of image generation in diffusion models \citep{Henderson-et-al-2023, Carlini-et-al-2022, Carlini-et-al-2023}.  Even if the training falls under ``fair use,'' this may violate the moral rights of content creators in some cases \citep{Weidinger-et-al-2022}.

    More subtly, generative models\footnote{See Chapters 12 and 14--18 of \citet{Prince-2023-UDL}.} raise novel questions regarding AI and intellectual property. Can the output of a machine learning model (e.g., art, music, code, text) be copyrighted or patented? Is it morally acceptable or legal to fine-tune a model on a particular artist's work to reproduce that artist's style? IP law is one area that highlights how existing legislation was not created with machine learning models in mind. Although governments and courts may set precedents in the near future, these questions are still open at the time of writing.

\subsection{Automation bias and moral deskilling}

    As society relies more on AI systems, there is an increased risk of automation bias (i.e., expectations that the model outputs are correct because they are ``objective''). This leads to the view that quantitative methods are better than qualitative ones. However, as we shall see in section~\ref{sec:ethics_Values}, purportedly objective endeavors are rarely value-free.

    The sociological concept of deskilling refers to the redundancy and devaluation of skills in light of automation \citep{Braverman-1974}. For example, off-loading cognitive skills like memory onto technology may cause a decrease in our capacity to remember things.  Analogously, the automation of AI in morally-loaded decision-making may lead to a decrease in our moral abilities \citep{Vallor-2014}.  For example, in the context of war, the automation of weapons systems may lead to the dehumanization of victims of war \citep{Asaro-2012, Heyns-2017}. Similarly, care robots in elderly-, child-, or healthcare settings may reduce our ability to care for one another \citep{Vallor-2011}.

\subsection{Environmental impact}

    Training deep networks requires significant computational power and hence consumes a large amount of energy. \citet{Strubell-et-al-2019, Strubell-et-al-2020} estimate that training a transformer model with $213$ million parameters emitted around 284 tonnes of $CO_2$.\footnote{As a baseline, it is estimated that the average human is responsible for around $5$ tonnes of $CO_2$ per year, with individuals from major oil-producing countries responsible for three times this amount. See \href{https://ourworldindata.org/co2-emissions}{https://ourworldindata.org/co2-emissions}.} \citet{Luccioni-et-al-2022} have provided similar estimates for the emissions produced from training the BLOOM language model. Unfortunately, the increasing prevalence of closed, proprietary models means that we know nothing about their environmental impacts \citep{Luccioni-2023}.

\subsection{Employment and society}

    The history of technological innovation is a history of job displacement. In 2018, the McKinsey Global Institute estimated that AI may increase economic output by approximately US \$13 trillion by 2030, primarily from the substitution of labor by automation \citep{Bughin-et-al-2018}. Another study from the McKinsey Global Institute suggests that up to 30\% of the global workforce (10-800 million people) could have their jobs displaced due to AI between 2016 and 2030 \citep{Manyika-et-al-2017, Manyika-Sneader-2018}.

    However, forecasting is inherently difficult and although automation by AI may lead to short-term job losses, the concept of {\it technological unemployment} has been described as a ``temporary phase of maladjustment'' \citep{Keynes-2010}. This is because gains in wealth can offset gains in productivity by creating increased demand for products and services. In addition, new technologies can create new types of jobs.

    Even if automation doesn't lead to a net loss of overall employment in the long term, new social programs may be required in the short term. Therefore, regardless of whether one is optimistic \citep{Brynjolfsson-McAfee-2016, Danaher-2019}, neutral \citep{Metcalf-et-al-2016, Calo-2018, Frey-2019}, or pessimistic \citep{Frey-Osborne-2017} about the possibility of unemployment in light of AI, it is clear that society will be changed significantly.

\subsection{Concentration of power}

    As deep networks increase in size, there is a corresponding increase in the amount of data and computing power required to train these models. In this regard, smaller companies and start-ups may not be able to compete with large, established tech companies. This may give rise to a feedback loop whereby the power and wealth become increasingly concentrated in the hands of a small number of corporations. A recent study finds an increasing discrepancy between publications at major AI venues by large tech firms and ``elite'' universities versus mid- or lower-tier universities \citep{Ahmed-Wahed-2020}. In many views, such a concentration of wealth and power is incompatible with just distributions in society \citep{Rawls-1971}.

    This has led to calls to democratize AI by making it possible for everyone to create such systems \citep{Fei-Fei-Li-2018, Knight-2018, Kratsios-2019, Riedl-2020}. Such a process requires making deep learning technologies more widely available and easier to use via open source and open science so that more people can benefit from them. This reduces barriers to entry and increases access to AI while cutting down costs, ensuring model accuracy, and increasing participation and inclusion \citep{Ahmed-et-al-2020}.

\section{Case study}

    We now describe a case study that speaks to many of the issues that we have discussed in this chapter.  In 2018, the popular media reported on a controversial facial analysis model---dubbed ``gaydar AI'' \citep{Wang-Kosinski-2018}---with sensationalist headlines like {\it AI Can Tell If You're Gay: Artificial Intelligence Predicts Sexuality From One Photo with Startling Accuracy} \citep{Ahmed-2017}; {\it A Frightening AI Can Determine Whether a Person Is Gay With 91 Percent Accuracy} \citep{Matsakis-2017}; and {\it Artificial Intelligence System Can Tell If You're Gay} \citep{Fernandez-2017}.

    There are a number of problems with this work.  First, the training dataset was highly biased and unrepresentative, being comprised mostly of Caucasian images. Second, modeling and validation are also questionable, given the fluidity of gender and sexuality. Third,  the most obvious use case for such a model is the targeted discrimination and persecution of LGBTQ+ individuals in countries where queerness is criminalized. Fourth, with regard to transparency, explainability, and value alignment more generally, the ``gaydar'' model appears to pick up on spurious correlations due to patterns in grooming, presentation, and lifestyle rather than facial structure, as the authors claimed \citep{Arcas-et-al-2018}. Fifth, with regard to  data privacy, questions arise regarding the ethics of scraping ``public'' photos and sexual orientation labels from a dating website. Finally, with regard to scientific communication, the researchers communicated their results in a way that was sure to generate headlines: even the title of the paper is an overstatement of the model's abilities: {\it Deep Neural Networks Can Detect Sexual Orientation from Faces}. (They cannot.)

    It should also be apparent that a facial-analysis model for determining sexual orientation does {\it nothing} whatsoever to benefit the LGBTQ+ community.  If it is to benefit society,  the most important question is whether a particular study, experiment, model, application, or technology serves the interests of the community to which it pertains.

\section{The value-free ideal of science}\label{sec:ethics_Values}

    This chapter has enumerated a number of ways that the objectives of AI systems can unintentionally, or through misuse, diverge from the values of humanity.   We now argue that scientists are not neutral actors; their values inevitably impinge on their work.

    Perhaps this is surprising.  There is a broad belief that science is---or ought to be---objective.  This is codified by the {\em value-free ideal of science}.  Many would argue that machine learning is objective because algorithms are just mathematics.  However, analogous to algorithmic bias (section~\ref{sec:ethics_Bias}), there are four stages at which the values of AI practitioners can affect their work \citep{Reiss-Sprenger-2017}:

    \begin{enumerate}
    \setlength \itemsep{-0.3em}
            \item The choice of research problem.
            \item Gathering evidence related to a research problem. 
            \item Accepting a scientific hypothesis as an answer to a problem. 
            \item Applying the results of scientific research. 
    \end{enumerate}

    It is perhaps uncontroversial that values play a significant role in the first and last of these stages.  The initial selection of research problems and the choice of subsequent applications are influenced by the interests of scientists, institutions, and funding agencies.   However, the value-free ideal of science prescribes minimizing the influence of moral, personal, social, political, and cultural values on the intervening scientific process. This idea presupposes the {\it value-neutrality thesis}, which suggests that scientists can (at least in principle) attend to stages (2) and (3) without making these value judgments.

    However, whether intentional or not, values are embedded in machine learning research. Most of these values would be classed as {\em epistemic} (e.g., performance, generalization, building on past work, efficiency, novelty).  But deciding the set of values is itself a value-laden decision; few papers explicitly discuss societal need, and fewer still discuss potential negative impacts \citep{Birhane-et-al-2022}.  Philosophers of science have questioned whether the value-free ideal of science is attainable or desirable. For example, \citet{Longino-1990, Longino-1996} argues that these epistemic values are not {\it purely} epistemic. \citet{Kitcher-2011a, Kitcher-2011b} argues that scientists don't typically care about {\it truth} itself; instead, they pursue truths relevant to their goals and interests.

    Machine learning depends on inductive inference and is hence prone to inductive risk. Models are only constrained at the training data points, and the curse of dimensionality means this is a tiny proportion of the input space; outputs can always be wrong, regardless of how much data we use to train the model.  It follows that choosing to accept or reject a model prediction requires a value judgment: that the risks if we are wrong in acceptance are lower than the risks if we are wrong in rejection.

    Hence, the use of inductive inference implies that machine learning models are deeply value-laden \citep{Johnson-2022}. In fact, if they were not, they would have no application: it is precisely {\it because} they are value-laden that they are useful. Thus, accepting that algorithms are used for ranking, sorting, filtering, recommending, categorizing, labeling, predicting, etc., in the real world implies that these processes will have real-world effects. As machine learning systems become increasingly commercialized and applied, they become more entrenched in the things we care about.

    These insights have implications for researchers who believe that algorithms are somehow more objective than human decision-makers (and, therefore, ought to replace human decision-makers in areas where we think objectivity matters).

\section{Responsible AI research as a collective action problem} \label{sec:ethics_CAP}

    It is easy to defer responsibility. Students and professionals who read this chapter might think their work is so far removed from the real world or a small part of a larger machine that their actions could not make a difference. However, this is a mistake. Researchers often have a choice about the projects to which they devote their time, the companies or institutions for which they work, the knowledge they seek, the social and intellectual circles in which they interact, and the way they communicate.

    Doing the right thing, whatever that may comprise, often takes the form of a social dilemma; the best outcomes depend upon cooperation, although it isn't necessarily in any individual's interest to cooperate: responsible AI research is a collective action problem.

\subsection{Scientific communication}

    One positive step is to communicate responsibly. Misinformation spreads faster and persists more readily than the truth in many types of social networks \citep{LaCroix-et-al-2021, Ceylan-et-al-2023}. As such, it is important not to overstate machine learning systems' abilities (see case study above) and to avoid misleading anthropomorphism. It is also important to be aware of the potential for the misapplication of machine learning techniques.  For example, pseudoscientific practices like phrenology and physiognomy have found a surprising resurgence in AI \citep{Stark-Hutson-2022}.

\subsection{Diversity and heterogeneity}

    A second positive step is to encourage diversity.  When social groups are homogeneous (composed mainly of similar members) or homophilous (comprising members that tend to associate with similar others), the dominant group tends to have its conventions recapitulated and stabilized \citep{Oconnor-Bruner-2019}. One way to mitigate systems of oppression is to ensure that diverse views are considered. This might be achieved through equity, diversity, inclusion, and accessibility initiatives (at an institutional level), participatory and community-based approaches to research (at the research level), and increased awareness of social, political, and moral issues (at an individual level).

    The theory of {\it standpoint epistemology} \citep{Harding-1986} suggests that knowledge is socially situated (i.e., depends on one's social position in society).  Homogeneity in tech circles can give rise to biased tech \citep{Noble-2018, Eubanks-2018, Benjamin-2019, Broussard-2023}. Lack of diversity implies that the perspectives of the individuals who create these technologies will seep into the datasets, algorithms, and code as the default perspective. \citet{Broussard-2023} argues that because much technology is developed by able-bodied, white, cisgender, American men, that technology is {\it optimized for} able-bodied, white, cisgender, American men, the perspective of whom is taken as the status quo. Ensuring technologies benefit historically marginalized communities requires researchers to understand the needs, wants, and perspectives of those communities \citep{Birhane-et-al-2022-Participatory-AI}. {\em Design justice} and participatory- and community-based approaches to AI research contend that the communities affected by technologies should be actively involved in their design \citep{Costanza-Chock-2020}.

\section{Ways forward}\label{sec:ethics_four_points}

    It is undeniable that AI will radically change society for better or worse. However, optimistic visions of a future Utopian society driven by AI should be met with caution and a healthy dose of critical reflection. Many of the touted benefits of AI are beneficial only in certain contexts and only to a subset of society. For example, \citet{Green-2019} highlights that one project developed using AI to enhance police accountability and alternatives to incarceration and another developed to increase security through predictive policing are both advertised as ``AI for Social Good.''  Assigning this label is a value judgment that lacks any grounding principles; one community's good is another's harm.

    When considering the potential for emerging technologies to benefit society, it is necessary to reflect on whether those benefits will be equally or equitably distributed.  It is often assumed that the most technologically advanced solution is the best one---so-called {\it technochauvinism} \citep{Broussard-2018}. However, many social issues arise from underlying social problems and do not warrant technological solutions.

    Some common themes emerged throughout this chapter, and we would like to impress four key points upon the reader:

\begin{enumerate}
\setlength \itemsep{-0.0em}
\item {\bf Research in machine learning cannot avoid ethics.} Historically, researchers could focus on fundamental aspects of their work in a controlled laboratory setting. However, this luxury is dwindling due to the vast economic incentives to commercialize AI and the degree to which academic work is funded by industry \cite[see][]{Abdalla-Abdalla-2021}; even theoretical studies may have social impacts, so researchers must engage with the social and ethical dimensions of their work.
           
\item{\bf Even purely technical decisions can be value-laden.} There is still a widely-held view that AI is fundamentally just mathematics and, therefore, it is ``objective,'' and ethics are irrelevant. This assumption is not true when we consider the creation of AI systems or their deployment.

\item {\bf We should question the structures within which AI work takes place.}  Much research on AI ethics focuses on specific situations rather than questioning the larger social structures within which AI will be deployed. For example, there is considerable interest in ensuring algorithmic fairness, but it may not always be possible to instantiate conceptions of fairness, justice, or equity within extant social and political structures. Therefore, technology is inherently political. 

\item {\bf Social and ethical problems don't necessarily require technical solutions.} Many potential ethical problems surrounding AI technologies are primarily social and structural, so technical innovation alone cannot solve these problems; if scientists are to effect positive change with new technology, they must take a political and moral position.
\end{enumerate}

Where does this leave the average scientist? Perhaps with the following imperative: it is necessary to reflect upon the moral and social dimensions of one's work. This might require actively engaging those communities that are likely to be most affected by new technologies, thus cultivating relationships between researchers and communities and empowering those communities. Likewise, it might involve engagement with the literature beyond one's own discipline.  For philosophical questions, the {\it Stanford Encyclopedia of Philosophy} is an invaluable resource. Interdisciplinary conferences are also useful in this regard. Leading work is published at both the Conference on Fairness, Accountability, and Transparency (FAccT) and the Conference on AI and Society (AIES).

\section{Summary}

This chapter considered the ethical implications of deep learning and AI. The value alignment problem is the task of ensuring that the objectives of AI systems are aligned with human objectives.  Bias, explainability, artificial moral agency, and other topics can be viewed through this lens.   AI can be intentionally misused, and this chapter detailed some ways this can happen.  Progress in AI has further implications in areas as diverse as IP law and climate change.   

Ethical AI is a collective action problem, and the chapter concludes with an appeal to scientists to consider the moral and ethical implications of their work. Every ethical issue is not within the control of every individual computer scientist.  However, this does not imply that researchers have no responsibility whatsoever to consider---and mitigate where they can---the potential for misuse of the systems they create.

\section*{Problems}
\small
\singlespacing

\paragraph{{\bf Problem 21.1}}\label{prob:ethics_value_alignment}
It was suggested that the most common specification of the value alignment problem for AI is ``the problem of ensuring that the values of AI systems are aligned with the values of humanity.'' Discuss the ways in which this statement of the problem is underspecified. \\ {\em Discussion Resource}: \cite{LaCroix2023}.

\phantom{a}

\paragraph{{\bf Problem 21.2}}\label{prob:goodharts_law}
Goodhart's law states that ``when a measure becomes a target, it ceases to be a good measure.'' Consider how this law might be reformulated to apply to value alignment for artificial intelligence, given that the loss function is a mere proxy for our true objectives.    % Asnwer: In AI, the loss function for a model is always a proxy for our true objectives, and when a proxy becomes the optimization target, it ceases to be a good proxy.

\phantom{a}

\paragraph{{\bf Problem 21.3}} \label{prob:ethics_biases}
Suppose a university uses data from past students to build models for predicting ``student success,'' where those models can support informed changes in policies and practices. Consider how biases might affect each of the four stages of the development and deployment of this model. \\ {\em Discussion Resource}: \citet{Fazelpour-Danks-2021}.

\phantom{a}

\paragraph{{\bf Problem 21.4}}  \label{prob:ethics_transparency}
We might think of functional transparency, structural transparency, and run transparency as orthogonal. Provide an example of how an increase in one form of transparency may not lead to a concomitant increase in another form of transparency. \\ {\em Discussion Resource}: \citet{Creel-2020}.

\phantom{a}

\paragraph{{\bf Problem 21.5}} \label{prob:ethics_responsibility}
If a computer scientist writes a research paper on AI or pushes code to a public repository, do you consider them responsible for future misuse of their work?

\phantom{a}

\paragraph{{\bf Problem 21.6}} \label{prob:ethics_militarization}
To what extent do you think the militarization of AI is inevitable?

\phantom{a}

\paragraph{{\bf Problem 21.7}} \label{prob:ethics_open}
In light of the possible misuse of AI highlighted in section~\ref{sec:ethics_Misuse}, make arguments both for and against the open-source culture of research in deep learning.

\phantom{a}

\paragraph{{\bf Problem 21.8}} \label{prob:privacy}
Some have suggested that personal data is a source of power for those who own it. Discuss the ways personal data is valuable to companies that utilize deep learning and consider the claim that losses to privacy are experienced collectively rather than individually. \\ {\em Discussion Resource}: \citet{Veliz-2020}. 

\phantom{a}

\paragraph{{\bf Problem 21.9}} \label{prob:ethics_ip}
What are the implications of generative AI for the creative industries?  How do you think IP laws should be modified to cope with this new development?

\phantom{a}

\paragraph{{\bf Problem 21.10}}\label{prob:forecasting}
A good forecast must (i) be specific enough to know when it is wrong, (ii) account for possible cognitive biases, and (iii) allow for rationally updating beliefs. Consider any claim in the recent media about future AI and discuss whether it satisfies these criteria.
\\ {\em Discussion Resource}: \cite{Tetlock-Gardner-2016}.

\phantom{a}

\paragraph{{\bf Problem 21.11}} \label{prob:democratize_ai}
Some critics have argued that calls to democratize AI have focused too heavily on the {\it participatory} aspects of democracy, which can increase risks of errors in collective perception, reasoning, and agency, leading to morally-bad outcomes. Reflect on each of the following: {\it What} aspects of AI should be democratized? {\it Why} should AI be democratized? {\it How} should AI be democratized? \\ {\em Discussion Resource}: \cite{Himmelreich-2022}.

\phantom{a}

\paragraph{{\bf Problem 21.12}}\label{prob:ethics_letter}
In March 2023, the Future of Life Institute published a letter, ``Pause Giant AI Experiments,'' in which they called on all AI labs to immediately pause for at least six months the training of AI systems more powerful than GPT-4.   Discuss the motivations of the authors in writing this letter, the public reaction, and the implications of such a pause.  Relate this episode to the view that AI ethics can be considered a collective action problem (section~\ref{sec:ethics_CAP}). \\ {\em Discussion Resource}: \citet{Gebru-et-al-2023}.

\phantom{a}

\paragraph{{\bf Problem 21.13}} \label{prob:ethics_four_points}
Discuss the merits of the four points in section~\ref{sec:ethics_four_points}.  Do you agree with them?

\newpage
\singlespacing
\bibliographystyle{apalikelike}
\bibliography{Biblio}

\end{document}